# Cross-Language Assessment of Mathematical Capability of ChatGPT

Gargi Sathe[1*], Aneesh Shamraj[1*], Aditya Surve[1*], Nahush Patil[1*], and Kumkum Saxena[2*]

[1]Thadomal Shahani Engineering College, Mumbai, India
{gargiss35,aneesh.shamraj,ssadityaa2002,patilnahush41}@gmail.com
[2]Thadomal Shahani Engineering College, Information Technology Department, Mumbai, India
kumkum.saxena@thadomal.org

**Abstract.** This paper presents an evaluation of the mathematical capability of ChatGPT across diverse languages like Hindi, Gujarati, and Marathi. ChatGPT, based on GPT-3.5 by OpenAI, has garnered significant attention for its natural language understanding and generation abilities. However, its performance in solving mathematical problems across multiple natural languages remains a comparatively unexplored area, especially in regional Indian languages. In this paper, we explore those capabilities as well as using chain-of-thought prompting to figure out if it increases the accuracy of responses as much as it does in the English language and provide insights into the current limitations.

**Keywords:** Generative Pre-Trained Transformer, Linguistics, Tokenization

## 1 Introduction

ChatGPT(Generative Pretrained Transformer) is considered a breakthrough in the domain of Natural Language Processing(NLP). Developed by OpenAI, it was publicly released on 30th November 2022 was all over the internet, and reached 1 million users in just 5 days. ChatGPT has an arsenal of capabilities including, essay writing, generating code, language translation, etc.

GPT 3.5 follows a strict four-stage training process to harness its language generation capabilities. The first step is pre-training in which the model understands the basics of language from a vast corpus of data. Supervised fine-tuning follows which improves responsiveness, while reward modeling evaluates and ensures response quality. Reinforcement learning optimizes the model and then tokenization converts text into input sequences for the decoder-only architecture of transformers [1-2].

In the Technical Report written by OpenAI on GPT-4, it was mentioned that GPT 3.5 gets scores corresponding to the 70th percentile on SAT math questions and the 25th percentile on GRE Quantitative questions [3-4]. However, we have observed that ChatGPT struggles in solving mathematical problems. After several months of utilizing ChatGPT for daily computational tasks, we observed discrepancies in its performance, leading us to question its accuracy.

*All authors have contributed equally in the process of this research paper.

Here's an example:

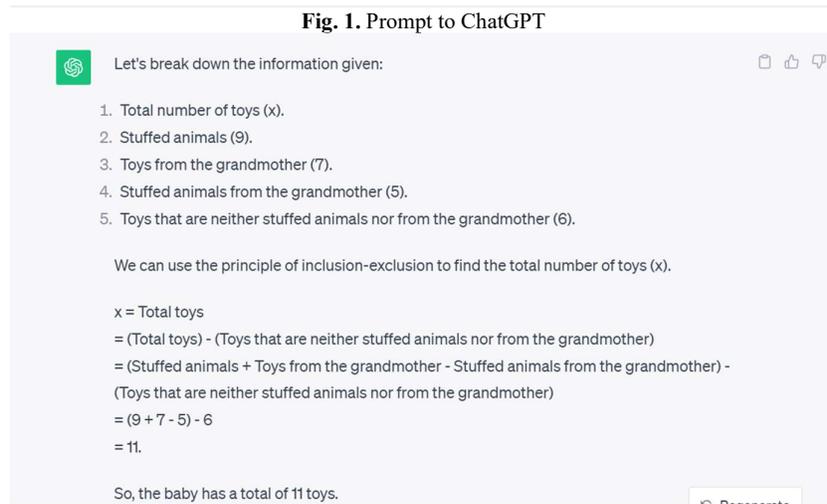

**Fig. 1.** Prompt to ChatGPT

**Fig. 2.** Example of ChatGPT making an error in the formula

When prompted, this set theory question, it understands the question well and makes an equation out of it, but at the last step, uses the wrong formula. The expected answer was 17 but the output given was 11.

Here's another example:

$$EOQ = \sqrt{\frac{2 \cdot 9,000 \cdot 100}{2.40}}$$
$$EOQ = \sqrt{375,000}$$
$$EOQ \approx 612.37$$

**Fig. 3.** Example of ChatGPT making a calculation error

In this specific part of the solution provided, ChatGPT makes an error in calculating the square root of the equation. The expected answer was 866.025.

## 1.1    Linguistic Challenges:

Mathematics is one of the most important subjects in the curriculum across India which is tested in schools, colleges, and various entrance examinations. These examinations, conducted by the respective state education departments for Secondary School Certificate (SSC), are offered in multiple languages including Marathi, Hindi, and Gujarati, etc. ChatGPT is primarily designed to understand and generate text in English, but it also supports at least 95 natural languages [5]. This gave us the basis to analyze the responses of ChatGPT in above stated three languages.

## 2    Methodology

Mathematical inaccuracies refer to the error between the actual value and the calculated value. We have tried ChatGPT on several mathematical word problems. These questions are taken from Secondary School Certificate Examinations conducted by the Maharashtra State Board (Board), Graduate Record Examination (GRE) reference book 5LB by Manhattan Prep, and the Maharashtra Common Entrance Test, (MHTCET) for engineering. We are considering dataset contamination to be a probable issue; therefore, we have carefully selected the MHTCET and Board questions from the examination year 2023[6].

We translated the questions using Google Translate and then gave them as input to ChatGPT only after human verification. If it gives the wrong answer, we took the following steps from our side to try and improve the performance:

- Rewrite the prompt using Romanization instead of the conventional Devnagari script.
- Regeneration of the answer

In terms of mathematical inaccuracies, we have considered the following questions to be answered:

- Does the generated response match the answer?
- Are the equations correctly produced in the response?
- Has it been able to comprehend the problem?

Taking into account the spatial limitations of this research paper, we will be considering the same two examples from our dataset to showcase the ability in the language English and the inability in non-English languages.

## 2.1    English Language Capabilities

English was the main language used to train ChatGPT and we have used it as a baseline for our evaluations to compare it with responses in other languages. Out of 25 questions of the dataset, English achieved 19 accurate responses.

**Question**:

The sum of the squares of five consecutive natural numbers is 1455. Find the numbers.

**Fig. 4.** First question given to ChatGPT in English

**Solution**:

Let's denote the five consecutive natural numbers as n, n+1, n+2, n+3, and n+4. The sum of their squares is given by the equation:

$$n^2 + (n+1)^2 + (n+2)^2 + (n+3)^2 + (n+4)^2 = 1455$$

Now, let's solve this equation. Expanding and simplifying each term, we get:

$$5n^2 + 20n + 30 = 1455$$

Combining like terms:

$$5n^2 + 20n - 1425 = 0$$

Now, we can factor the quadratic equation:

$$n^2 + 4n - 285 = 0$$

Factoring, we find:

$$(n-15)(n+19) = 0$$

This gives two possible solutions for n: $n = 15$ or $n = -19$. Since we are looking for natural numbers, we discard the negative solution.

So, the consecutive natural numbers are 15, 16, 17, 18, and 19.

**Fig. 5.** ChatGPT correctly finds a solution in English.

**Question**:

In triangle ABC, segment AP is a median. If BC = 18, AB^2 + AC^2 = 260, Find AP.

**Fig. 6.** Second question given to ChatGPT in English

**Solution**:

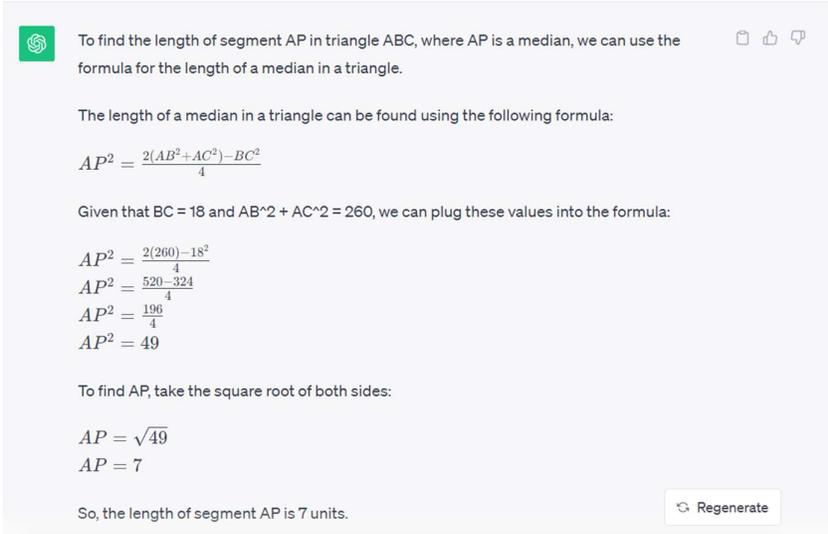

To find the length of segment AP in triangle ABC, where AP is a median, we can use the formula for the length of a median in a triangle.

The length of a median in a triangle can be found using the following formula:

$$AP^2 = \frac{2(AB^2 + AC^2) - BC^2}{4}$$

Given that BC = 18 and AB^2 + AC^2 = 260, we can plug these values into the formula:

$$AP^2 = \frac{2(260) - 18^2}{4}$$
$$AP^2 = \frac{520 - 324}{4}$$
$$AP^2 = \frac{196}{4}$$
$$AP^2 = 49$$

To find AP, take the square root of both sides:

$$AP = \sqrt{49}$$
$$AP = 7$$

So, the length of segment AP is 7 units.

**Fig. 7.** ChatGPT correctly finding a solution in English

As seen in the above images, we received correct responses when the questions were given in English for these two questions.

## 2.2 Hindi Language Capabilities
Hindi is spoken by about 41% of the Indian population [7]. Out of 25 math questions of the dataset, for Hindi language, it achieved 10 accurate responses.

**Question**:

N पांच क्रमागत प्राकृतिक संख्याओं के वर्गों का योग 1455 है।
संख्याएँ ढूँढें।

**Fig. 8.** First question given to ChatGPT in Hindi

**Solution**:

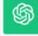

**Fig. 9.** ChatGPT not comprehending the question

As seen here, ChatGPT misinterpreted the question, and instead of finding the natural numbers, it took the first 5 natural numbers' squares.

**Question**:

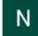

**Fig. 10.** Second question in Hindi

**Solution**:

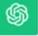

**Fig. 11.** ChatGPT not misinterpreting the question

As shown in Fig. 11, ChatGPT misinterpreted the question. It tried to use the Pythagorean Theorem even though the existence of a right-angled triangle wasn't mentioned instead of using the Apollonius Theorem.

## 2.3 Marathi Language Capabilities

Marathi is the third most spoken language in India with about 7% of the total population of India speaking Marathi [7]. It is the official language of Maharashtra and has a significant number of native speakers. Out of 25 math questions given to ChatGPT, it got 9 correct.

**Question**:

सलग पाच नैसर्गिक संख्यांच्या वर्गांची बेरीज 1455 आहे. संख्यांचा क्रम शोधा

**Fig. 12.** First question given to ChatGPT in Marathi

**Solution**:

43^2 = 1849

44^2 = 1936

45^2 = 2025

आपल्याला 45 च्या वर्गनि 1455 लाभले आहे.

**Fig. 13.** ChatGPT wrongly calculating output in Marathi

In the given question it is asked to find the sequence of 5 natural numbers whose sum of squares is 1455. Instead of using algebra to solve it, it just lists out the squares to 45 and deviates from the initial question. Further, it concludes that 1455 is the square of 45 even though it itself had given the square of 45 as 2025.

**Question**:

ABC त्रिकोणामध्ये, AP हा मध्यक आहे. BC = 18, AB^2 + AC^2 = 260 असल्यास, AP शोधा.

**Fig. 14.** Second question given to ChatGPT in Marathi

**Solution**:

ABC त्रिकोणामध्ये, AP हा मध्यक आहे. BC = 18, AB^2 + AC^2 = 260 असल्यास, AP शोधा.

AP शोधण्याचा एक उपाय त्रिकोणाच्या वायरे आणि परिपधने उपयोग करण्‌ा आहे. त्रिकोणाच्या वायरे म्हणजे बीसी आणि एपी ची नेमकी नेमकी शिर्षक वाचते तर तीन कोणाच्या संबंधाने,

बीसी^2 = AB^2 + AC^2

18^2 = AB^2 + AC^2

324 = AB^2 + AC^2

AP यांची नेमकी त्रिकोणाच्या शिर्षिकाच्या नेमक्या च्या नेमक्या सारखी आहे:

**Fig. 15.** ChatGPT making an error in the formula.

In the question above, it forms the wrong equation and hence fails to calculate the correct answer.

## 2.4    Gujarati Language Capabilities

Gujarati is the sixth most widely spoken language in India [7]. It is primarily spoken in the Indian state of Gujarat and some regions of the neighboring states. Out of the 25 math questions presented to it, it got 7 right.

**Question**:

સતત પાંચ પ્રાકૃતિક સંખ્યાઓના વર્ગોનો સરવાળો 1455 છે. નંબરો શોધો.

તમે પાંચ પ્રાકૃતિક સંખ

**Fig. 16.** ChatGPT in Gujarati not being able to understand the question.

**Solution**:

ચાલો સ્ટેપ બાય સ્ટેપ વિચારીએ

ખુશ રહો, આવું કરવાનું સ્ટેપ-બાય-સ્ટેપ પ્રક્રિયા:

1. **સરવાળીનો વર્ગ કે ઝાડ સંખ્યા મેળવો:**
   • 1455 ને સરવાળો વર્ગ કે ઝાડ સંખ્યા બનાવવામાં મદદ કરે છે. સામાન્યત: √1455 = 38.09.
2. **વર્ગમાંથી વિલીન્દ્ઃ**
   • સરવાળા વર્ગનો વિલીન્દ્ર સંખ્યા 38.09 છે. આ સંખ્યાનું વિલીન્દ્ર વર્ગમાંથી વાગવું છે.
3. **સંખ્યાનો વર્ગ:**
   • આપેલા વર્ગનું વર્ગ 38.09 નો મૂળછે. આપેલા સંખ્યાનું વર્ગ હોય છે 6.17.
4. **સરવાળીની દોઢી મૂલ્ય શોધો:**
   • વર્ગ કે ઝાડ સંખ્યાની દોઢી મૂલ્ય અને વર્ગની દોઢી મૂલ્ય સાથે સામાન્ય છે. તાહલે, 1455 ની દોઢી મૂલ્ય છે 38.
5. **સંખ્યાને પ્રમેય કરો:**
   • આપેલી પ્રક્રિયા અનુસરીને, સંખ્યા 1455 સોલ્યુશન છે.

   આ પ્રક્રિયાને સમામ કરવાનો પરિણામ છે 1455 સંખ્યા છે અને તેનું સરવાળું અને વર્ગછું સરવાળું વર્ગ છે.

**Fig. 17.** Upon regeneration of the same question.

As you can see, the first response to the question from ChatGPT is 'You are five natural numbers.'
Solution:
Even after using chain of thought prompting in Gujarati, it still produces an unintelligible response hence producing an undesired result.

**Question**:

ત્રિકોણ ABC માં, સેગમેન્ટ AP એ મધ્યક છે. જો BC = 18, AB^2 + AC^2 = 260, AP શોધો.

**Fig. 18.** Question two in Gujarati language.

**Solution**:

આપેલી સમસ્યાને પ્યાથગોરસ થીઓરમની ઉપયોગ કરીને નેમાંકના પ્રમાણાત્મક રીતે સોલ્વ કરો:

1. આપેલી માહિતીની પરિણામિક જાણકારી:
   • BC = 18
   • AB^2 + AC^2 = 260
2. BC, AB, અને AC ને પ્યાથગોરસ થીઓરમની ઉપયોગ કરીને સંજ્ઞાન લો:
   • BC^2 = AB^2 + AC^2
3. એટલે,
   18^2 = AB^2 + AC^2

**Fig. 19.** ChatGPT making an error in the concept.

The question demanded the concept of 'Properties of Triangles'. After engaging in the Pythagorean Theorem, it gets stuck in a loop problem and keeps repeating the same step.

It was also observed that ChatGPT frequently struggles to generate complete responses in all non-English languages. We had to regenerate responses often when it was asked to solve questions in Hindi, Marathi, and Gujarati.

In all the non-English languages, the process of comprehending the mathematical question seems to be drastically different. In English, it first understands the question, builds logical equations on it, and then solves them. On the other hand, for Hindi, Marathi, and Gujarati, it tries to give responses around the numbers in the prompt rather than form logical equations.

## 2.5    Prompt Modification

Strategic prompt engineering, such as adding "chain of thought" prompts, can greatly enhance the model's ability to solve mathematical word problems correctly. We used "Let's think step by step" as our chain of thought prompt. This type of prompting guides the model to think step-by-step and provide the reasoning behind the math, rather than just generating text [8].

In the context of the English language, an analysis comprising 25 questions revealed that 10 questions required prompt modification, and from those, 4 questions exhibited correctness while the other 6 still gave the wrong answer. However, it is crucial to note that the benefits derived from chain-of-thought prompting appear to be language-specific. When applied to languages other than English, the analysis demonstrated minimal to no discernible improvement in performance. This highlights the need for language-specific considerations and adaptations when implementing strategic prompt engineering techniques.

## 3. Conclusion

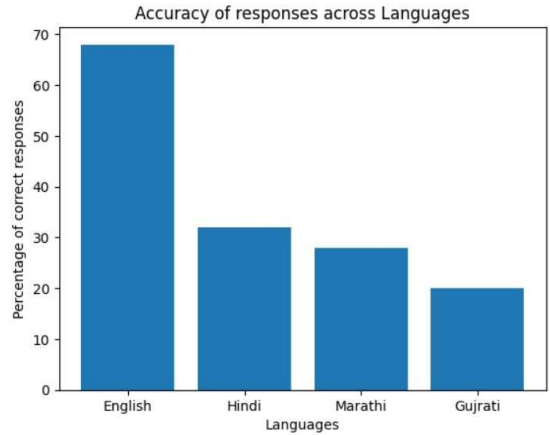

**Fig. 20.** Comparison Bar Graph of analysis.

As shown in Fig. 20, the results in Hindi, Marathi, and Gujarati had a substantial drop in accuracy with accuracy in Hindi being at 32%, Marathi at 28%, and Gujarati at 20%.

It is found that GPT-3.5 performs worse when the questions are in languages like Hindi, Marathi, and Gujarati as compared to English. The results in English are found to be similar to the ones found in the GPT-4 Technical Report having an accuracy of 68% with an increase in accuracy by using chain-of-thought prompting.

## 4 Future Scope

GPT-3.5 has great capabilities to solve mathematics and reasoning problems as compared to its other LLM counterparts. However, there is still uncertainty and unreliability, regarding its solving abilities in English and other languages. We observed that it can form equations out of word problems and it can solve standalone arithmetic equations, but in some cases, it is unable to do these two processes together. There seems to be a gap between understanding and solving due to which it often gives incorrect answers. In some instances, it gave different answers on regenerating the response. When given multiple choice questions, and has given the wrong option as the answer, and when given the correct option as prompt, it re-calculates the answer and vice-versa. Large Language Models are a big step in the field of AI and have made various daily life tasks easy. However the working of LLMs and how it computes the output is still a mystery. We might have better clarity once we understand what is going on inside the black box.